\PassOptionsToPackage{usenames}{color}
\documentclass[11pt]{article}
\usepackage{hyperref}	
\hypersetup{
	hidelinks,
	urlcolor=blue
}
\usepackage{relsize} 
\usepackage{acl2013}
\usepackage[boxed]{algorithm2e}
\usepackage[small,bf,skip=5pt]{caption}
\usepackage{sidecap} 

\usepackage{titlesec}
\titleformat*{\subparagraph}{\itshape}
\titlespacing{\subparagraph}{%
  1em}{
  0pt}{
  1em}

\usepackage{lingmacros}

\usepackage{enumitem} 
\setitemize{noitemsep,topsep=0em}
\setenumerate{noitemsep,leftmargin=0em,itemindent=13pt,topsep=0em}

\usepackage{textcomp}

\usepackage{listings}

\usepackage{amsmath,amsfonts,eucal,amsbsy,amsthm,amsopn,amssymb}

\usepackage{txfonts}
\usepackage[scaled]{beramono}
\usepackage[T1]{fontenc}

\usepackage{latexsym}

\usepackage{multirow}
\usepackage{booktabs} 
\usepackage{multicol}
\usepackage{footnote}

\usepackage{url}
\usepackage[usenames]{color}
\usepackage[normalem]{ulem} 
\usepackage{colortbl}
\usepackage{graphicx}
\usepackage{subfig}

\usepackage{color}
\usepackage{bm}
\definecolor{orange}{rgb}{1,0.5,0}
\definecolor{mdgreen}{rgb}{0,0.6,0}
\definecolor{dkblue}{rgb}{0,0,0.5}
\definecolor{dkgray}{rgb}{0.3,0.3,0.3}
\definecolor{slate}{rgb}{0.25,0.25,0.4}
\definecolor{gray}{rgb}{0.5,0.5,0.5}
\definecolor{ltgray}{rgb}{0.7,0.7,0.7}
\definecolor{purple}{rgb}{0.7,0,1.0}
\definecolor{lavender}{rgb}{0.65,0.55,1.0}

\newcommand{\Sref}[1]{\S\ref{#1}}
\newcommand{\fref}[1]{figure~\ref{#1}}
\newcommand{\ffref}[2]{figures~\ref{#1} and~\ref{#2}}

\newcommand{\tref}[1]{table~\ref{#1}}
\newcommand{\ttref}[2]{tables~\ref{#1} and~\ref{#2}}
\newcommand{\Tref}[1]{Table~\ref{#1}}
\newcommand{\aref}[1]{algorithm~\ref{#1}}

\newcommand{\fnref}[1]{footnote~\ref{#1}}

\makeatletter
\renewcommand{\paragraph}{%
  \@startsection{paragraph}{4}%
  {\z@}{.2ex \@plus 1ex \@minus .2ex}{-1em}%
  {\normalfont\normalsize\bfseries}%
}
\makeatother

\newcommand{\gfl}[1]{
\mbox{\textsmaller{\texttt{#1}}}}	
\newcommand{\dataset}[1]{\mbox{\textsc{#1}}}	

\newcommand{\finalversion}[1]{#1}

\newcommand{\considercutting}[1]{#1}
\newcommand{\longversion}[1]{} 

\title{A Framework for (Under)specifying Dependency Syntax without~Overloading Annotators}

\finalversion{\author{\begin{tabular}{@{}>{\centering}p{.7\textwidth}@{}>{\centering}p{.32\textwidth}@{}}
Nathan Schneider\thanks{\ \ Corresponding author: \href{mailto:nschneid@cs.cmu.edu}{\tt nschneid@cs.cmu.edu}}\hspace{10pt}Brendan O'Connor\hspace{7pt}Naomi Saphra\\
David Bamman\hspace{7pt}Manaal Faruqui\hspace{7pt}Noah A. Smith\hspace{7pt}Chris Dyer
& 
\rm School of Computer Science \\ Carnegie Mellon University 
\end{tabular}\\[10pt]
\bf Jason Baldridge\hspace{30pt}\rm Department of Linguistics, The University of Texas at Austin
}}

\date{}

\begin{document}
\maketitle

\begin{abstract}
We introduce a framework for lightweight dependency syntax annotation.
Our formalism builds upon the typical representation for unlabeled dependencies, 
permitting a simple notation and annotation workflow. 
Moreover, the formalism encourages annotators to underspecify parts of the syntax 
if doing so would streamline the annotation process.
We demonstrate the efficacy of this annotation on three languages 
and develop algorithms to evaluate and compare underspecified annotations.
{\bf This paper is an expanded version of \cite{fudg-law} 
containing additional technical details.}

\end{abstract}


\section{Introduction}

Computational representations for natural language syntax are borne of
competing design considerations. When designing such representations, there may be a tradeoff between parsimony and expressiveness. 
A range of \emph{linguistic theories} attract support due to differing
purposes and aesthetic principles \cite[\textit{inter alia}]{chomsky-57,tesniere-59,hudson-84,sgall-86,melcuk-88}. 
Formalisms concerned with tractable computation may care chiefly about \emph{learnability} or \emph{parsing efficiency} \cite{shieber-92,sleator-93,kuhlmann-06}.
Further considerations may include \emph{psychological} and \emph{evolutionary plausibility} \cite{croft-01,tomasello-03,steels-11,fossum-12}, 
integration with other representations such as semantics \cite{steedman-00,bergen-05}, 
or suitability for particular applications (e.g., translation).

Here we elevate \emph{ease of annotation} as a primary design concern for a syntactic annotation formalism.
Currently, a lack of annotated data is a huge bottleneck for robust NLP, standing in the way of parsers 
for social media text \cite{foster-11} and many low-resourced languages (to name two examples).
Traditional syntactic annotation projects like the Penn Treebank \cite{ptb} or Prague Dependency Treebank \cite{hajic-98}
require highly trained annotators and huge amounts of effort.
Lowering the cost of annotation, by making it easier and more accessible,
could
greatly facilitate robust NLP in new languages and genres,
and allow researchers to more easily experiment with new tasks and problems.

\begin{figure*}\small
\begin{multicols}{2}
{\smaller\color{blue}{\sf Found the scarriest mystery door in my school\,.\,I'M SO CURIOUS D:}}
\vspace{-3pt}

\begin{verse}
\gfl{Found** < (the scarriest mystery door*)}\\ 
\gfl{Found < in < (my > school)}\\
\gfl{I'M** < (SO > CURIOUS)}\\
\gfl{D:**}\\
\gfl{my = I'M}
\end{verse}

\noindent {\smaller\color{blue}{\sf thers still like 1 1/2 hours till Biebs bday here :P}}
\vspace{-3pt}

\begin{verse}
\gfl{thers** < still}\\  
\gfl{thers < ((1 1/2) > hours < till < (Biebs > bday))}\\
\gfl{(thers like 1 1/2 hours)}\\
\gfl{thers < here}\\ 
\gfl{:P**}
\end{verse}
\end{multicols}
\vspace{-6pt}

\caption{Two tweets with example GFL annotations. 
(The formalism and notation are described in \Sref{sec:gfl}.)}\label{fig:tweetex}
\end{figure*}

To that end, we design and test new, lightweight methodologies for syntactic annotation. 
We propose a \emph{formalism}, \textbf{Fragmentary Unlabeled
  Dependency Grammar} (FUDG) for unlabeled dependency syntax that addresses some of the most glaring deficiencies 
of basic unlabeled dependencies (\Sref{sec:motivation}), with little added burden on annotators. 
FUDG requires minimal theoretical commitments, 
and can be supplemented with a language- or project-specific style guide (we provide a brief one for English).
To facilitate efficient annotation, we contribute a 
simple ASCII markup language, \textbf{Graph Fragment Language} (GFL;
\Sref{sec:gfl}), that allows annotations to be authored using any text
editor. We release tools that make it
possible to validate, normalize, and visualize GFL annotations.\footnote{\textsmaller{\url{https://github.com/brendano/gfl_syntax/}}\label{fn:resources}}

An important characteristic of our framework is annotator flexibility. 
The formalism supports this by allowing \emph{underspecification} of structural portions that 
are unclear or unnecessary for the purposes of a project. Fully leveraging this power requires new algorithms for 
evaluation, e.g., of inter-annotator agreement, where annotations are
partial; such algorithms are presented in
\Sref{sec:eval}.\footnote{Parsing algorithms are left for future
  work.\label{fn:no-parsing-here}}

Finally, small-scale case studies (\S\ref{sec:casestudies}) apply our framework
(formalism, notation, and evaluations)  
to syntactically annotate social web text in English, news in Malagasy, and dialogues in Kinyarwanda\longversion{, two languages spoken in Africa from the Austronesian and Bantu families, respectively}.

\section{A Dependency Grammar for Annotation}\label{sec:motivation}

Although dependency-based approaches to syntax play a major role in computational linguistics,
the nature of dependency representations is far from uniform.
Exemplifying one end of the spectrum is the Prague Dependency Treebank,
which articulates an elaborate dependency-based syntactic theory in a rich, multi-tiered formalism \cite{hajic-98,bohmova-03}.
On the opposite end of the spectrum are the structures used in
dependency parsing research which organize all the tokens of a
sentence into a tree, sometimes with category labels on the edges \cite{kubler-09}.
Insofar as they reflect a theory of syntax, these \textbf{vanilla dependency grammars} provide a highly reductionist 
view of structure---indeed, parses used to train and evaluate dependency parses are often simplifications 
of Prague-style parses, or else converted from constituent treebanks.

In addition to the binary dependency links of vanilla dependency representations, we offer three devices 
to capture certain linguistic phenomena more straightforwardly:\footnote{Some of this is inspired by the conventions of Reed-Kellogg \emph{sentence diagramming},
a graphical dependency annotation system for English pedagogy \cite{reed-77,kolln-94,florey-06}.
\label{sentdiag}}

\begin{enumerate}

\item We make explicit the meaningful lexical units over which syntactic structure is represented. 
Our approach (a)~allows punctuation and other extraneous tokens to be excluded 
so as not to distract from the essential structure; 
and (b)~permits tokens to be grouped into shallow multiword lexical units.\footnote{The Stanford representation 
supports a limited notion of multiword expressions \cite{de_marneffe-08}. For simplicity, our formalism treats multiwords 
as unanalyzed (syntactically opaque) wholes, though some multiword expressions may have syntactic descriptions \cite{baldwin-10}.}

\item Coordination is problematic to represent with unlabeled dependencies 
due to its non-binary nature.
A coordinating conjunction typically joins multiple expressions (conjuncts) with equal status, 
and other expressions may relate to the compound structure as a unit.
There are several different conventions for forcing coordinate structures into a head-modifier straightjacket \cite{nivre-05,de_marneffe-08,marecek-13}.
Conjuncts, coordinators, and shared dependents can be distinguished with edge labels; 
we equivalently use a special notation, permitting the coordinate structure to be automatically 
transformed with any of the existing conventions.\footnote{\newcite{tesniere-59} and \newcite{hudson-84} similarly use special structures for coordination \cite{schneider-98,sangati-09}.}

\item Following \newcite{tesniere-59}, our formalism offers a simple facility to express anaphora-antecedent relations 
(a subset of semantic relationships) that are salient in particular syntactic phenomena such as relative clauses, appositives, 
and \textsc{wh}-expressions.

\end{enumerate}

\subsection{Underspecification}\label{sec:underspec}

Our desire to facilitate lightweight annotation scenarios requires us 
to abandon the expectation that syntactic informants provide a 
complete parse for every sentence. 
On one hand, an annotator may be \emph{uncertain} about the appropriate parse  
due to lack of expertise, insufficiently mature annotation conventions, 
or actual ambiguity in the sentence.
On the other hand, annotators may be \emph{indifferent} to certain phenomena. This can happen for a variety of reasons, including:
\begin{itemize}
\item Some projects may only need annotations of specific constructions. For example, building a semantic resource for events may require annotation 
of syntactic verb-argument relations, but not internal noun phrase structure.
\item As a project matures, it may be more useful to annotate only infrequent lexical items.
\item Semisupervised learning from partial annotations may be sufficient to learn complete parsers \cite{hwa-99,clark-06}.
\item Beginning annotators may wish to focus on easily understood syntactic phenomena.
\item Different members of a project may wish to specialize in different syntactic phenomena,
reducing training cost and cognitive load.
\end{itemize}
Rather than treating annotations as invalid unless and until they are complete trees, 
we formally represent and reason about partial parse structures.
Annotators produce \textbf{annotations}, which encode constraints on the (inferred) \textbf{analysis}, the parse structure, of a sentence. We say that a valid annotation \textbf{supports} (is compatible with) one or more \textbf{analyses}. Both annotations and analyses are represented as graphs~(the graph representation is described below in~\S\ref{sec:graph-encoding}). We require that the directed edges in an \emph{analysis} graph 
must form a tree over all the lexical items in the sentence.\footnote{While some linguistic phenomena (e.g., relative clauses, control constructions) can be represented using non-tree structures, we find that being able to alert annotators when they inadvertently violate the tree constraint is more useful than the expressive flexibility.} Less stringent well-formedness 
constraints on the \emph{annotation} graph leave room for underspecification.

Briefly, an annotation can be underspecified in two ways: (a)~an expression may not be attached 
to any parent, indicating it might depend on any nondescendant in a full analysis---this is useful for 
annotating sentences piece by piece; and 
(b)~multiple expressions may be grouped together in a \textbf{fudge expression} (\S\ref{sec:means}), 
a constraint that the elements form a connected subgraph in the full analysis while leaving the 
precise nature of that subgraph indeterminate---this is useful for marking relationships 
between chunks (possibly constituents)\longversion{ of a sentence}. 

\subsection{A Formalism, not a Theory}

Our framework for dependency grammar annotation is a syntactic \emph{formalism}, but it is not sufficiently comprehensive to constitute a \emph{theory} of syntax. 
Though it standardizes the basic treatment of a few basic phenomena, simplicity of the formalism requires us to be conservative about making such extensions. 
Therefore, just as with simpler formalisms, language- and project-specific conventions will have to be developed for specific linguistic phenomena. 
By embracing underspecified annotation, however, our formalism aims to encourage efficient corpus coverage in a nascent annotation project, without forcing annotators to make premature decisions.

\section{Syntactic Formalism and GFL}\label{sec:gfl}

In our framework, a syntactic \textbf{annotation} of a sentence follows an extended dependency formalism based on the desiderata enumerated in the previous section. We call our formalism \textbf{Fragmentary Unlabeled Dependency Grammar} (\textbf{FUDG}).

To make it simple to create FUDG annotations with a text editor,
we provide a plain-text dependency notation called \textbf{Graph Fragment Language} (\textbf{GFL}). 
Fragments of the FUDG graph---nodes and dependencies linking them---are encoded in this language; taken together, 
these fragments describe the annotation in its entirety.
The ordering of GFL fragments, and of tokens within each fragment, is of no formal consequence.
Since the underlying FUDG representation is transparently related to GFL constructions, GFL notation will be introduced alongside the discussion of each kind
of FUDG node.\footnote{%
In principle, FUDG annotations could be created with an alternative mechanism such as a GUI,
as in \newcite{hajic-01}.}

\subsection{Tokens}

We expect a tokenized string, such as a sentence or short
message.
The provided tokenization is respected in the annotation.  For human readability, GFL fragments refer to tokens as strings (rather than offsets), so all tokens that participate in an annotation must be unambiguous in the input.\footnote{If a word is repeated within the sentence, it must be indexed in the input string
in order to be referred to from a fragment. In our notation, successive instances of the same word are suffixed with \gfl{\texttildelow 1}, \gfl{\texttildelow 2}, \gfl{\texttildelow 3}, etc.
Punctuation and other tokens omitted from an annotation do not need to
be indexed.}  A token may be referenced multiple times in the annotation.

\subsection{Graph Encoding}
\label{sec:graph-encoding}

\paragraph{Directed arcs.} As in other dependency formalisms, \textbf{dependency arcs} are directed links 
indicating the syntactic headedness relationship between pairs of nodes. 
\longversion{

 }In GFL, directed arcs are indicated with angle brackets pointing from the dependent to its head, 
as in \gfl{black > cat} or (equivalently) \gfl{cat < black}. Multiple arcs can be 
chained together: \gfl{the > cat < black < jet} describes three arcs. Parentheses help group 
portions of a chain: \gfl{(the > cat < black < jet) > likes < fish} (the structure \gfl{black < jet > likes}, 
in which \gfl{jet} appears to have two heads, is disallowed). Note that another encoding for this structure would be 
to place the contents of the parentheses and the chain \gfl{cat > likes < fish} on separate lines. 
Curly braces can be used to list multiple dependents of the same head: \gfl{\{cat fish\} > likes}. 

\begin{figure*}[ht]\small\centering
\vspace{-0.2in}
\includegraphics[width=6in]{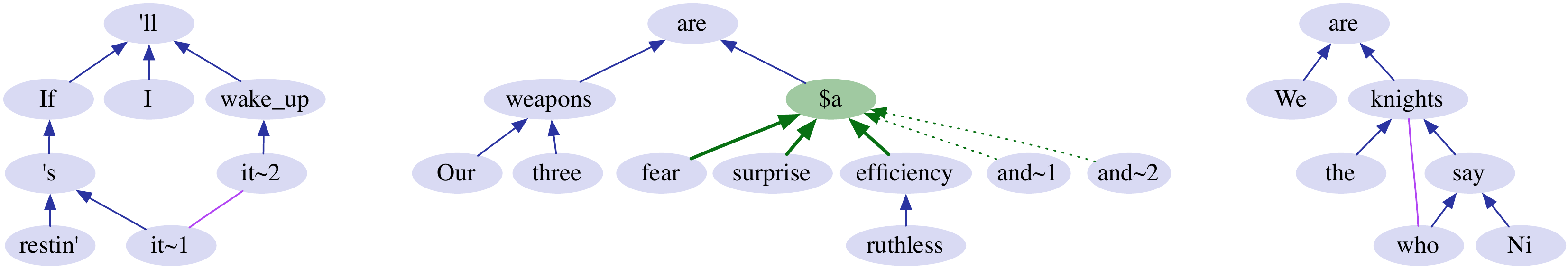}
\caption{FUDG graphs corresponding to the examples in \fref{fig:gflex}. The two special kinds of directed edges are for attaching conjuncts (bolded) and their coordinators (dotted) in a coordinate structure. 
Anaphoric links are undirected. The root node of each sentence is omitted.}\label{fig:fudgex}
\end{figure*}

\paragraph{Anaphoric links.} These undirected links join coreferent anaphora to each other and to their antecedent(s). 
In English this includes personal pronouns, relative pronouns (\textit{who}, \textit{which}, \textit{that}), and 
anaphoric \textit{do} and \textit{so} (\textit{Leo loves Ulla and \textbf{so does} Max}). 
This introduces a bit of semantics into our annotation, though at present we do not attempt to mark non-anaphoric coreference. 
It also allows a more satisfying treatment of appositives and relative clauses than would be possible from just the directed tree 
(the third example in \ffref{fig:fudgex}{fig:gflex}).

\paragraph{Lexical nodes.} Whereas in vanilla dependency grammar syntactic links are between pairs of \textbf{token nodes}, 
FUDG abstracts away from the individual tokens in the input.
The lowest level of a FUDG annotation consists of \textbf{lexical
  nodes}, i.e., lexical item occurrences.\longversion{\footnote{In principle, 
the lexical nodes could normalize the spelling of the input token, 
or could even point to entries in a lexicon such as WordNet \cite{wordnet}. 
For our annotation, however, the name of the lexical node will simply be the concatenation of the input tokens it represents.}} 
Every token node maps to 0 or 1 lexical nodes (punctuation\longversion{ tokens}, for instance, can be ignored). 

A \textbf{multiword} is a lexical node incorporating more than one input token and is atomic (does not contain internal structure). 
A multiword node may group any subset of input tokens; 
this allows for multiword expressions which are not necessarily contiguous in the sentence 
(e.g., the verb-particle construction \textit{make up} in \textit{make the story up}). 
GFL notates multiwords with square brackets, e.g., \gfl{[break a leg]}.

\paragraph{Coordination nodes.} Coordinate structures require at least two kinds of dependents: 
\textbf{coordinators} (i.e., lexical nodes for coordinating conjunctions---at least one per coordination node) and 
\textbf{conjuncts} (heads of the conjoined subgraphs---at least one per coordination node). 
The GFL annotation has three parts: a variable representing the node, a set of conjuncts, and 
a set of coordinator nodes. For instance, \gfl{\$a ::\ \{[peanut butter] honey\}\ ::\ \{and\}} 
(\textit{peanut butter and honey}) can be embedded within a phrase via the coordination node variable \gfl{\$a};
\textit{a [fresh [[peanut butter] and honey] sandwich] snack} 
would be formed with \gfl{\{fresh \$a\} > sandwich > snack < a}.
A graphical example of coordination can be seen in \fref{fig:fudgex}---note the bolded conjunct edges and 
the dotted coordinator edges.

If the conjoined phrase as a whole takes modifiers, these are attached to the coordination node with regular directed arcs.
For example, in \textit{Sam really adores kittens and abhors puppies.}, the shared subject \textit{Sam} and adverb \textit{really} 
attach to the entire conjoined phrase. In GFL:
\vspace{-10pt}\begin{verse}
 \gfl{\$a ::\ \{adores abhors\}\ ::\ \{and\}}\\
 \gfl{Sam > \$a < really}\\
 \gfl{adores < kittens}\\
 \gfl{abhors < puppies}
\end{verse}
\vspace{-10pt}
\begin{figure*}[ht]
\small\centering
\vspace{-0.2in}
\includegraphics[height=.9in]{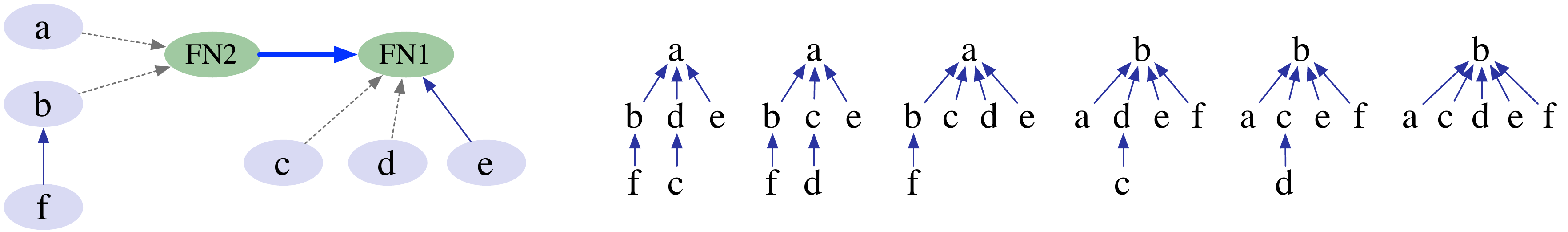}
\caption{\textit{Left:} An annotation graph with 2 fudge nodes and 6 lexical nodes; 
it can be encoded with GFL fragments \gfl{((a b)* c d) < e} and \gfl{b < f}.
\textit{Right:} All of its supported analyses: $prom(\mathcal{A})=6$. 
$com(\mathcal{A}) = 1-\frac{\log{6}}{\log{7^5}} = .816$.
}\label{fig:fudgenodes}
\end{figure*}

\paragraph{Root node.} This is a special top-level node used to indicate that a graph fragment constitutes 
a standalone utterance or a discourse connective. For an input with multiple utterances,  
the head of each should be designated with \gfl{**} to indicate that it attaches to the root.

\subsection{Means of Underspecification}
\label{sec:means}

As discussed in \Sref{sec:underspec}, our framework distinguishes \emph{annotations} from full syntactic \emph{analyses}. 
With respect to dependency structure (directed edges), the former may underspecify the latter, 
allowing the annotator to commit only to a partial analysis.

For an annotation $\mathcal{A}$, we define $\textit{support}(\mathcal{A})$ to be the set of full analyses compatible with that annotation. 
A full analysis is required to be a directed rooted tree over all lexical nodes in the annotation.
An annotation is \emph{valid} if its support is non-empty.

FUDG has two mechanisms for dependency underspecification:
unattached nodes and fudge nodes.

\paragraph{Unattached nodes.} For any node in an annotation, the annotator is free to simply leave it not attached to any head.
This is interpreted as allowing its head to be any other node
(including the root node), subject to the tree constraint. 
We call a node's possible heads its \textbf{supported parents}.
Formally, for an unattached node $v$ in annotation $\mathcal{A}$, 
$\textit{suppParents}_{\mathcal{A}}(v)=\textit{nodes}(\mathcal{A})\setminus (\{v\} \cup \textit{descendants}(v))$.

\paragraph{Fudge nodes.} Sometimes, however, it is desirable to represent a sort of skeletal structure without filling in all the details. 
A \textbf{fudge expression} (FE) asserts that a group of nodes (the expression's \textbf{members}) belong together in a \emph{connected subgraph}, 
while leaving the internal structure of that subgraph unspecified.\footnote{This underspecification semantics is, to the best of our knowledge, novel, 
though it has been proposed that connected dependency subgraphs (known as \emph{catenae}) are of theoretical importance in syntax \cite{osborne-12}.} 
The notation for this is a list of two or more nodes within parentheses: 
an annotation for \textit{Few if any witches are friends with Maria.}\ might contain the FE \gfl{(Few if any)} so as to 
be compatible with the structures \gfl{Few < if < any}, \gfl{Few > if > any}, etc.---but \emph{not}, for instance, \gfl{Few > witches < any}.
In the FUDG graph, this is represented with a \textbf{fudge node} to which members are attached by special \textbf{member arcs}.
Fudge nodes may be linked to other nodes: the GFL fragment \gfl{(Few if any) > witches} is 
compatible with \gfl{(Few < if < any) > witches}, \gfl{(Few < (if > any)) > witches}, and so forth.

\subparagraph{Properties.} Let $\mathbf{f}$ be a fudge expression. 
From the connected subgraph definition and the tree constraint on analyses, it follows that:
\begin{itemize}
	\item Exactly 1 member of $\mathbf{f}$ must, in any compatible analysis, have a parent that is not a member of $\mathbf{f}$.
	Call this node the \textbf{top} of the fudge expression, denoted $f^*$. $f^*$ dominates all other members of $\mathbf{f}$; it can be considered $\mathbf{f}$'s ``internal head.''
	\item $\mathbf{f}$ does not necessarily form a full subtree. Any of its members may have dependents that are not themselves members of the fudge expression. 
	(Such dependencies can be specified in additional GFL fragments.)
\end{itemize}

\subparagraph{Top designation.} A single member of a fudge expression may optionally be designated as its top (internal head). 
This is specified with an asterisk: \gfl{(Few* if any) > witches} indicates that \textit{Few} must attach to \textit{witches} 
and also dominate both \textit{if} and \textit{any}. In the FUDG graph, this is represented with a special \textbf{top arc} 
as depicted in bold in \fref{fig:fudgenodes}.

\subparagraph{Nesting.} One fudge expression may nest within another, e.g. \gfl{(Few (if any)) > witches}; 
the word analyzed as attaching to \gfl{witches} might be \gfl{Few} or whichever of \gfl{(if any)} heads the other. A nested 
fudge expression can be designated as top: \gfl{(Vanishingly few (if any)*)}\longversion{ thus implies consistency with either 
\gfl{(Vanishingly few if*)} or \gfl{(Vanishingly few any*)}}.

\subparagraph{Modifiers.} An arc attaching a node to a fudge expression as a whole asserts that the external node 
should modify the \emph{top} of the fudge expression (whether or not that top is designated in the annotation). 
For instance, two of the interpretations of \textit{British left waffles on Falklands} would be preserved by specifying
\gfl{British > left} and \gfl{(left waffles) < on < Falklands}. Analyses \gfl{British > left < waffles < on < Falklands} and 
\gfl{(British > left < on < Falklands) > waffles} would be excluded because the preposition does not attach to the head of \gfl{(left waffles)}.\footnote{Not all attachment ambiguities can be precisely encoded in FUDG. For instance, there is no way to forbid an attachment to a word 
that lies along the path between the possible heads. 
\considercutting{The best that can be done given a sentence like 
\textit{They conspired to defenestrate themselves on Tuesday.}\ is \gfl{They > conspired < to < defenestrate < themselves} and 
\gfl{(conspired* to defenestrate (on < Tuesday))}.}
}

\subparagraph{Multiple membership.} A node may be a member of multiple fudge expressions, 
or a member of an FE while attached to some other node via an explicit arc.
Each connected component of the FUDG graph is therefore a polytree (not necessarily a tree). 
The annotation graph minus all member edges of fudge nodes and all (undirected) anaphoric links must be a directed tree or forest.

\subparagraph{Enumerating supported parents.} Fudge expressions complicate the procedure for listing a node's supported parents 
(see above). Consider an FE $\mathbf{f}$ having some member $v$. $v$ might be the top of $\mathbf{f}$ 
(unless some other node is so designated), in which case anything the fudge node can attach to is a potential parent of $v$. 
If some node other than $v$ might be the top of $\mathbf{f}$, then $v$'s head could be any member of $\mathbf{f}$. 
Below (\Sref{sec:promis}) we develop an algorithm for enumerating supported parents for any annotation graph node.

\begin{figure}\small
\vspace{0.1in}
\href{http://youtu.be/e6Lq771TVm4?t=51s}{\color{blue}\gfl{If it\texttildelow1 \textquotesingle s restin\textquotesingle\ I \textquotesingle ll wake it\texttildelow2 up .}} \\
\phantom{\quad}\gfl{If < (it\texttildelow1 > \textquotesingle s < restin\textquotesingle)}\\
\phantom{\quad}\gfl{I > \textquotesingle ll < [wake up] < it\texttildelow2}\\
\phantom{\quad}\gfl{If > \textquotesingle ll**}\\
\phantom{\quad}\gfl{it\texttildelow1 = it\texttildelow2}\\
\href{http://youtu.be/Tym0MObFpTI?t=1m10s}{\color{blue}\gfl{Our three weapons are fear and\texttildelow1 surprise and\texttildelow2} \\
\gfl{ruthless efficiency \ldots}} \\
\phantom{\quad}\gfl{\{Our three\} > weapons > are < \$a} \\
\phantom{\quad}\gfl{\$a ::\ \{fear surprise efficiency\}\ ::\ \{and\texttildelow1 and\texttildelow2\}} \\
\phantom{\quad}\gfl{ruthless > efficiency}\\
\href{http://youtu.be/0e2kaQqxmQ0?t=44s}{\color{blue}\gfl{We are the knights who say \ldots\ Ni !}}\\
\phantom{\quad}\gfl{We > are < knights < the}\\
\phantom{\quad}\gfl{knights < (who > say < Ni)}\\
\phantom{\quad}\gfl{who = knights}
\caption{GFL for the FUDG graphs in
  \fref{fig:fudgex}. \label{fig:gflex}
  }
\end{figure}

\section{Annotation Evaluation Measures}\label{sec:eval}

Quantifying inter-annotator agreement on a common dataset is an important tool for reasoning about individual annotators 
and the overall difficulty of the annotation task (influenced by the nature of the data, the annotation process, and the 
training\slash{}guidelines\slash{}reference materials available to annotators). If there is a well-defined set of decisions that need to be 
made for each sentence, the two annotators' choices can be compared directly. However, when the task allows for a great deal of latitude---as 
in our case, where a syntactic annotation may be full or partial---the comparison becomes more difficult. 
We formalize techniques for comparing two annotations in our formalism, keeping in mind that the structure may be underspecified in different 
parts or to different degrees. We must further address:

\begin{itemize}
\item \textbf{Annotation efficiency}, quantified in terms of annotator productivity (tokens per hour).

\item The \textbf{amount of information} in an underspecified annotation. 
Intuitively, an annotation that flirts with many full analyses conveys less syntactic information 
than one which supports few analyses.  
We define an annotation's \textbf{promiscuity} to be the number of full analyses it supports,
and develop an algorithm to compute it (\S\ref{sec:promis}). 

\item \textbf{Inter-annotator agreement} between two partial annotations.
Our measures for dependency structure agreement (\S\ref{sec:iaa}) incorporate the notion of promiscuity.
\end{itemize}

We test these evaluations on our pilot annotation data in the case studies (\Sref{sec:casestudies}).

\subsection{Promiscuity vs.~Commitment}\label{sec:promis}

Given a FUDG annotation of a sentence, we quantify the extent to which it underspecifies 
the full structure by counting the number of analyses that are compatible with the constraints in the annotation. 
We call this number the \textbf{promiscuity} of the annotation.
Each analysis tree is rooted with the root node and must span all lexical nodes.\footnote{\label{fn:simplification}This measure assumes a fixed lexical analysis (set of lexical nodes) and does not consider anaphoric links. 
Coordinate structures are simplified into ordinary dependencies, with coordinate phrases headed by the coordinator's lexical node. 
If a coordination node has multiple coordinators, one is arbitrarily chosen as the head and the others as its dependents.} 

A na\"{i}ve algorithm for computing promiscuity would be to enumerate
all directed spanning trees over the lexical nodes, 
and then check each of them for compatibility with the annotation. 
But this quickly becomes intractable: for $n$ nodes, one of which is designated as the root, there are $n^{n-2}$ spanning trees. 
However, we can filter out edges that are known to be incompatible with the annotation \emph{before} searching for spanning trees. 
Our ``upward-downward'' method for constructing a graph of \textbf{supported edges} first enumerates a set of candidate top nodes for every 
fudge expression (\aref{alg:upward}), then uses that information to infer a set of supported parents for every node (\aref{alg:downward}).
The supported edge graph then consists of vertices $\textit{lexnodes}(\mathcal{A}) \cup \{\textsc{root}\}$ and edges 
$\bigcup_{v \in \textit{lexnodes}(\mathcal{A})}{\{(v\rightarrow v') \
  \forall\  v' \in \textit{suppParents}_{\mathcal{A}}(v)\}}$. 
From this graph we then extract all directed spanning trees using the algorithm of \newcite{uno-96}.
If some lexical node has no supported parents, this reflects conflicting constraints in the annotation, 
and no spanning tree will be found.

\begin{lstlisting}[float,frame=single,caption={Identify the possible top nodes for each fudge node, 
traversing the annotation graph fragments
bottom-up.},label=alg:upward]
def upward(F):
  for n in sorted(F.nodes, key=lambda v: v.height):
    if n.isFudge:
      n.topcandidates = set()
      for (c,e) in n.childedges:
        if e=='top':
          n.topcandidates = tc(c)
          break    # there can be only one top edge
        elif e=='member':
          n.topcandidates |= tc(c)

def tc(n):
  return set(n.topcandidates) if n.isFudge else {n}
\end{lstlisting}
\begin{lstlisting}[float,frame=single,caption={Identify nodes' possible parents in an analysis
by traversing the annotation graph top-down. ``Firm'' nodes are all non-fudge nodes. Set operators: 
\textsmaller{\texttt{|}}~(union), \textsmaller{\texttt{\&}}~(intersection), \textsmaller{\texttt{-}}~(difference).},label=alg:downward]
def downward(G):
  for n in sorted(G.nodes, key=lambda v: v.depth):
    if not n.isRoot:
      n.suppParents = G.firmNodes-{n}-n.descendants
      for (p,e) in n.parentedges:
        if p.isFudge:
          if n in p.members:
            cands = set()
            if p.topcandidates & tc(n):
              # (top of) n might be the top of p
              cands |= p.suppParents
            if p.topcandidates!=tc(n):
              # (top of) n might not be the top of p
              siblings = p.members - {n}
              for sib in siblings:
                cands |= tc(sib)
            n.suppParents &= cands
          else: # external modifier of a fudge node
            n.suppParents &= p.topcandidates
        else: # explicit parent a lexical node or root
            n.suppParents &= {p}
\end{lstlisting}

One final step is needed to  ensure that non-member attachments to a fudge node, such as $e$ in \fref{fig:fudgenodes}, 
attach to the top node of the fudge expression. The structure \gfl{a < b < e} is \emph{not} supported by the annotation, though its individual edges are.
Such spanning trees must be filtered out post hoc. Alternatively, Kirchhoff's matrix tree theorem \cite{chaiken-78,smith-07,margoliash-10}
can be used to count all spanning trees in cubic time without enumerating them, and thus obtain an upper bound on promiscuity.\footnote{This 
would also allow efficient probabilistic inference over analyses given annotations, at least under common modeling assumptions.}
In \Sref{sec:casestudies} we use ``Exact'' promiscuity for annotations where enumerating all analyses is tractable, 
and ``Kirchhoff'' promiscuity for approximately measuring all annotations.

Promiscuity will tend to be higher for longer sentences. To control for this, we define a second quantity, the annotation's 
\textbf{commitment quotient} (commitment being the opposite of promiscuity), which normalizes for the number of possible spanning trees 
given the sentence length. The commitment quotient for an annotation of a sentence with $n-1$ lexical nodes and one root node is given by: 
\begin{equation*}
com(\mathcal{A}) = \frac{\log{n^{n-2}}-\log{prom(\mathcal{A})}}{\log{n^{n-2}}-\log{1}} = 1 - \frac{\log{prom(\mathcal{A})}}{\log{n^{n-2}}}
\end{equation*}
(the logs are to attenuate the dominance of the exponential term).
This will be 1 if only a single tree is supported by the annotation, and 0 if the annotation does not constrain the structure at all.
(If the constraints in the annotation are internally inconsistent, then promiscuity will be 0 and commitment undefined.)
In practice, there is a tradeoff between efficiency and commitment: more detailed annotations 
require more time. The value of minimizing promiscuity will therefore 
depend on the resources and goals of the annotation project.

\subsection{Inter-Annotator Agreement}\label{sec:iaa}

FUDG can encode flat groupings and coreference at the lexical level, as 
well as syntactic structure over lexical items. Inter-annotator agreement can be 
measured separately for each of these facets. 
Pilot annotator feedback indicated that our initial lexical-level guidelines were inadequate, 
so we focus here on measuring structural agreement pending further
clarification of the lexical conventions. 

Attachment accuracy, a standard measure for evaluating dependency parsers, 
cannot be computed between two FUDG annotations if either of them
underspecifies any part of the dependency structure.
One solution is to consider the intersection of supported full trees, in the spirit of our promiscuity measure. 
For annotations $\mathcal{A}_1$ and $\mathcal{A}_2$ of sentence $\mathbf{s}$, 
one annotation's supported analyses can be enumerated and then filtered
subject to the constraints of the other annotation. 
The tradeoff between inter-annotator compatibility and commitment can be accounted for by taking their product, 
i.e. $\textit{comPrec}(\mathcal{A}_1\mid\mathcal{A}_2) =
\textit{com}(\mathcal{A}_1)\frac{|\textit{supp}(\mathcal{A}_1) \cap
  \textit{supp}(\mathcal{A}_2)|}{|\textit{supp}(\mathcal{A}_1)|}$.

A limitation of this support-intersection approach is that if the two annotations are not compatible, 
the intersection will be empty. A more fine-grained approach is to decompose the comparison by lexical node: 
we generalize attachment accuracy with 
$\textit{softComPrec}(\mathcal{A}_1\mid\mathcal{A}_2) = \textit{com}(\mathcal{A}_1)\frac{\sum_{\ell \in \mathbf{s}}{\bigcap_{i \in \{1,2\}}\textit{suppParents}_{\mathcal{A}_i}(\ell)}}{\sum_{\ell \in \mathbf{s}}{\textit{suppParents}_{\mathcal{A}_1}(\ell)}}$,
taking advantage of the algorithms in the previous section to compute $\textit{com}(\cdot)$ and $\textit{suppParents}(\cdot)$. 
As lexical nodes may differ between the two annotations, a reconciliation step is required to compare the structures:
multiwords proposed in only one of the two annotations are converted to fudge expressions. 
Tokens annotated by neither annotator are ignored; excluding punctuation is standard practice in 
dependency evaluation \cite{buchholz-06}.
Like with the promiscuity measure, we simplify coordinate structures to ordinary dependencies (see \fnref{fn:simplification}).

\section{Case Studies}\label{sec:casestudies}

\subsection{Annotation Time}

\begin{table}
\centering\small
\begin{tabular}{lrr}
\bf Language & \bf Tokens & \bf Rate (tokens/hr) \\
English Tweets (partial) & 667 & 430\hspace{25pt} \\
English Tweets (full) & 388 & 250\hspace{25pt} \\
Malagasy & 4,184 & 47\hspace{25pt} \\
Kinyarwanda & 8,036 & 80\hspace{25pt} \\
\end{tabular}
\caption{Productivity estimates from pilot annotation project. All annotators were native speakers of English.}\label{tbl:productivity}
\vspace{-10pt}
\end{table}
To estimate annotation efficiency,
we performed a pilot annotation project consisting of
annotating several hundred English tweets, about 1,000 sentences in
Malagasy, and a further 1,000
sentences in Kinyarwanda.\footnote{Malagasy is a VOS Austronesian
language spoken by
  15 million people, mostly in Madagascar.  Kinyarwanda is an SVO Bantu
  language spoken by 12 million people mostly in Rwanda.  All
  annotations were done by native speakers of English. The
  Kinyarwanda and Malagasy annotators had basic proficiency in these
  languages.}
Table~\ref{tbl:productivity} summarizes the number of tokens annotated
and the effort required. For the two Twitter cases, the same annotator
was first permitted to do partial annotation (specifically, her
instructions were to leave unannotated all punctuation and any ``Twitter discourse''
markers that did not participate in syntactic relations)
of 100 tweets, and then spend the same amount of time doing a complete
annotation of all tokens. Although this is a very small study, the
results clearly suggest she was able to make much more rapid progress
when partial annotation was an option.\footnote{As a point of
comparison, during the Penn Treebank project, annotators corrected the
syntactic bracketings produced by a high-quality hand-written parser
(Fidditch) and achieved a rate of only 375 tokens/hour using a
specialized GUI interface \cite{ptb}.}

\begin{table*}[ht]
\centering\small
\begin{tabular}{crrrrrcrr@{ }r@{ }>{\smaller}r@{ }>{\smaller}r@{ }>{\smaller}r@{ }>{\smaller}rr}
&         &         & \bf Omitted & \bf Coordi- & \multicolumn{1}{c}{\bf Anaph}  & \bf             & \multicolumn{1}{c}{\bf Fudge} & \multicolumn{6}{c}{\bf $prom$ Histogram} & \multicolumn{1}{c}{\bf Mean}   \\
& \bf 1Ws & \bf MWs & \multicolumn{1}{c}{\bf Tokens}  & \multicolumn{1}{c}{\bf nations} & \multicolumn{1}{c}{\bf Links} & \bf Utterances & \multicolumn{1}{c}{\bf Nodes} & 1 & $>$1 & $\geq$10 & $\geq$$10^2$ & $\geq$$10^3$ & $\geq$$10^4$ & \multicolumn{1}{c}{\bf $com$} \\
\cmidrule{2-15}
\multicolumn{1}{l}{\dataset{Tweets}} & \multicolumn{8}{l}{\it 60 messages, 957 tokens} \\
A & 597 & 56 & 304 & 10 & 25 & $[146,152]$ & 23 & 43 & 17 & 11 & 5 & 4 & 2 & .96 \\
B & 644 & 47 & 266 & 8  &  8 & $[156,169]$ & 28 & 37 & 23 & 12 & 6 & 2 & 1 & .95 \\
\multicolumn{1}{l}{\dataset{Reviews}} & \multicolumn{8}{l}{\it 55 sentences, 778 tokens} \\
A & 609 & 33 & 136 & 23 & 30 & $[86,90]$   &  2 & 53 &  2 &  2 & 1 & 1 & 1 & 1.00 \\
C $\cap$ D & 643 & 19 & 116 & 17 & 13 & $[88,132]$ & 114 & 11 & 44 & 38 & 21 & 14 & 10 & .82 \\
T & 704 & --- & 74 & --- & --- & 62 & --- & 55 & 0 & 0 & 0 & 0 & 0 & 1\phantom{.00} \\
\end{tabular}
\caption{Measures of our annotation samples. Note that annotator ``D'' specialized in noun phrase--internal structure and personal pronoun anaphora, 
while annotator ``C'' specialized in verb phrase/clausal phenomena; $\textrm{C}\cap\textrm{D}$ denotes the combination of their annotation fragments. 
``T'' denotes our dependency conversion of the English Web Treebank parses. 
(The value 1.00 was rounded up from .9994.)}\label{tbl:measures}
\end{table*}

This pilot study helped us to identify linguistic phenomena 
warranting specific conventions: these include \textsc{wh}-expressions, comparatives, 
vocatives, discourse connectives, null copula constructions, and many others.
We documented these cases in a 20-page style guide for English,\footnote{Included with the data and software release (\fnref{fn:resources}).} 
which informed the subsequent pilot studies discussed below.

\subsection{Underspecification and Agreement}

With detailed guidelines in place, we then annotated
two small English data samples 
in order to study annotators' use of underspecification. 
The first sample is drawn from Owoputi~et~al.'s~\shortcite{owoputi-tr} 
Twitter part-of-speech corpus.\footnote{Specifically, the \dataset{Daily547} portion, 
downloaded from \textsmaller{\url{http://www.ark.cs.cmu.edu/TweetNLP/}}} 
The second is from the \dataset{Reviews} portion of the English Web Treebank \cite{webtb}, 
which contains Penn Treebank--style constituent parses. (Our annotators only saw the tokenized text.)
Both datasets are informal and conversational in nature, and are dominated by short messages/sentences. 
In spite of their brevity, many of the items were deemed to contain multiple ``utterances,'' which we define to include 
discourse connectives and emoticons (at best marginal parts of the syntax); 
utterance heads are marked with \gfl{**} in \fref{fig:tweetex}.

\Tref{tbl:measures} indicates the sizes of the two data samples, and gives statistics 
over the output of each annotator: total counts of single-word and multiword lexical nodes, 
tokens not represented by any lexical node, coordination nodes, anaphoric links, utterances,\footnote{Utterance counts are given as ranges, 
as any annotation fragment not explicitly headed by the root node may or may not form its own utterance.} 
and fudge nodes; as well as a histogram of promiscuity counts and the average of commitment quotients (see \Sref{sec:promis}).
For instance, the two sets of annotations obtained for the \dataset{Tweets} sample used underspecification 
in 17/60 and 23/60 tweets, respectively, though the promiscuity rarely exceeded 100 compatible trees per annotation.
Examples can be seen in \fref{fig:tweetex}, where annotator ``A'' marked only the noun phrase head for \textit{the scarriest mystery door}, 
opted not to choose a head within the quantity \textit{1~1/2},
and left ambiguous the attachment of the hedge \textit{like}.
The strong but not utter commitment to the dependency structure is reflected in the mean commitment quotients for this dataset, 
both of which exceed $0.95$.

\begin{table*}[ht]
\centering\small
\begin{tabular}{l rrlc rrr p{0.7em} rrrc rrr}
 & \multicolumn{7}{c}{\textbf{Exact} (exponential counting)} && \multicolumn{7}{c}{\textbf{Kirchhoff} ($O(n^3)$ counting)} \\
\cmidrule(lr){2-8}\cmidrule(lr){10-16}
 &  & \multicolumn{2}{c@{\ }}{$\textit{com}$} &  & \multicolumn{3}{c@{\ }}{$\textit{softComPrec}$} &
 &  & \multicolumn{2}{c@{\ }}{$\textit{com}$} &  & \multicolumn{3}{c@{\ }}{$\textit{softComPrec}$} \\
\cmidrule(lr){3-4}\cmidrule(lr){6-8}\cmidrule(lr){11-12}\cmidrule(lr){13-16}
 & \multicolumn{1}{c@{\ }}{$N$} & \multicolumn{1}{c}{$1$} & \multicolumn{1}{c}{$2$} & \multicolumn{1}{c}{$N_{|\cap|>0}$} 
 & \multicolumn{1}{c}{${1|2}$} & \multicolumn{1}{c}{${2|1}$} & \multicolumn{1}{c}{$F_1$} &
 & \multicolumn{1}{c@{\ }}{$N$} & \multicolumn{1}{c}{$1$} & \multicolumn{1}{c}{$2$} & \multicolumn{1}{c}{$N_{|\cap|>0}$} 
 & \multicolumn{1}{c}{${1|2}$} & \multicolumn{1}{c}{${2|1}$} & \multicolumn{1}{c}{$F_1$} \\
\cmidrule{2-16}
\multicolumn{1}{@{}l}{\dataset{Tweets}} & /60 &&&&&&&& /60  \\
\hspace{5pt}$\textrm{A}\sim\textrm{B}$ & 53 & .86 & \phantom{1}.93 & 22 &  .62 & .74 & .67 && 60 & .82 & .91 & 27 & .57 & .72 & .63 \\
\multicolumn{1}{@{}l}{\dataset{Reviews}} & /55 &&&&&&&& /55 \\
\hspace{5pt}$\textrm{A}\sim(\textrm{C}\cap\textrm{D})$ & 41 & .97 & \phantom{1}.80 & 27  & .68 & .48 & .53 && 55 & .95 & .76 & 30 & .64 & .40 & .50 \\
\hspace{5pt}$\textrm{A}\sim\textrm{T}$ & 50 & .95 & 1\phantom{.00} & 25 & .52 & .91 & .66 && 55 & .92 & 1\phantom{.00} & 26 & .48 & .91 & .63 \\
\hspace{5pt}$(\textrm{C}\cap\textrm{D})\sim\textrm{T}$ & 38 & .78 & 1.00 & 22 & .42 & .93 & .58 && 55 & .73 & 1.00 & 28 & .33 & .93 & .49 \\
\end{tabular}
\caption{Measures of inter-annotator agreement. 
Annotator labels are as in \tref{tbl:measures}.
Using exact measurement of promiscuity/commitment, the number $N$ of sentences whose trees can be enumerated for both annotations 
is limited.
Within this subset, $N_{|\cap|>0}$ counts sentences whose combined annotation supports at least one tree.
Per-annotator $\textit{com}$ (with lexical reconciliation) and inter-annotator
$\textit{softComPrec}$ are aggregated over sentences by arithmetic mean. \vspace{5pt}
}\label{tbl:iaa}
\end{table*}

Inter-annotator agreement (IAA) is quantified in \tref{tbl:iaa}. The row marked $\textrm{A}\sim\textrm{B}$, for instance, considers the 
agreement between annotator ``A'' and annotator ``B''. A handful of individual annotations are too underspecified for enumeration of 
all analyses to be tractable; sentences having this property for either of the two annotations are removed from consideration in the ``Exact'' 
measurements, as indicated by the $N$ column. Measuring IAA on the dependency structure requires a common set of 
lexical nodes, so a \textbf{lexical reconciliation} step ensures that (a)~any token used by either annotation is present in both, and 
(b)~no multiword node is present in only one annotation---this is solved by relaxing incompatible multiwords to fudge expressions
(which increases promiscuity). 
For \dataset{Tweets}, lexical reconciliation thus reduces the commitment averages for each annotation---to a greater extent for annotator ``A'' 
(.96 in \tref{tbl:measures} vs.~.86 in \tref{tbl:iaa}) because ``A'' marked more multiwords. 
An analysis fully compatible with both annotations exists for only 22/53 sentences; the finer-grained 
$\textit{softComPrec}$ measure (\Sref{sec:iaa}), however, offers insight into the balance between commitment and agreement.

The right half of \tref{tbl:iaa} uses Kirchhoff's matrix tree theorem (see \Sref{sec:promis}) to estimate the number of supported analyses 
for all sentences, including those that support too many analyses to enumerate. 
(The Kirchhoff count is an upper bound on promiscuity.) In general, the averages seen with the Kirchhoff estimates are comparable 
to the averages with the exact promiscuity\slash{}commitment measures.

Qualitatively, we observe three leading causes of incompatibilities (disagreements): 
obvious annotator mistakes (such as \textit{the} marked as a head); inconsistent handling of verbal auxiliaries; and 
uncertainty whether to attach expressions to a verb or the root node, as with \textit{here} in \fref{fig:tweetex}.\footnote{Another example: Some uses of conjunctions like 
\textit{and} and \textit{so} could be interpreted either as phrasal coordinators or as discourse connectives \nocite{pdtb-manual}(cf. The PDTB Research Group, 2007).}
Annotators noticed occasional ambiguous cases and attempted to encode the ambiguity with fudge expressions: 
\textit{again} in the tweet \textit{maybe put it off until you feel like \textasciitilde\ talking again ?}\ is one example.
More commonly, fudge expressions proved useful for syntactically difficult
constructions, such as those shown in \fref{fig:tweetex} as well as: \textit{such a good night}, 
\textit{asked what tribe I was from}, \textit{you two}, \textit{2~shy of breaking it}, 
\textit{a~\oldstylenums{\$}~13~/~day charge}, and \textit{the most awkward thing ever}.

\subsection{Annotator Specialization}

As an experiment in using underspecification for labor division,
two of the annotators of \dataset{Reviews} data were assigned specific linguistic phenomena to focus on. 
Annotator ``D'' was tasked with the internal structure of base noun phrases, including resolving the antecedents of personal pronouns. 
``C'' was asked to mark the remaining phenomena---i.e., utterance\slash{}clause\slash{}verb phrase structure---but 
to mark base noun phrases as fudge expressions, leaving their internal structure unspecified. Both annotators provided a full lexical analysis. 
For comparison, a third individual, ``A,'' annotated the same data in full. 
The three annotators worked completely independently. 

Of the results in \ttref{tbl:measures}{tbl:iaa}, the most notable difference between full and specialized annotation 
is that the combination of independent specialized annotations ($\textrm{C}\cap\textrm{D}$) 
produces somewhat higher promiscuity\slash{}lower commitment. 
This is unsurprising because annotators sometimes overlook relationships that fall under their specialty.\footnote{A more practical 
and less error-prone approach might be for specialists to work sequentially or collaboratively 
(rather than independently) on each sentence.}
Still, annotators reported that specialization made the task less burdensome, 
and the specialized annotations did prove complementary to each other.\footnote{In fact, for only 2 sentences did ``C'' and ``D'' have 
incompatible annotations, and both were due to simple mistakes that were then fixed in the combination.}

\subsection{Treebank Comparison}

Though the annotators in our study were native speakers well acquainted with representations of English syntax, 
we sought to quantify their agreement with the expert treebankers who created the English Web Treebank 
(the source of the \dataset{Reviews} sentences).
To convert the Treebank's constituent parses to dependencies, 
we used the PennConverter tool \cite{johansson-07}
with options 
chosen to emulate our annotation conventions,\footnote{Downloaded from \textsmaller{\url{http://nlp.cs.lth.se/software/treebank_converter/}} and run with \textsmaller{\texttt{-rightBranching=false -coordStructure=prague -prepAsHead=true -posAsHead=true -subAsHead=true -imAsHead=true -whAsHead=false}}} 
and then removed punctuation tokens.

Agreement with the converted treebank parses appears in the bottom two rows of \tref{tbl:iaa}. 
Because the Treebank commits to a single analysis, precision scores are quite lopsided. 
Most of its attachments are consistent with our annotations ($\textit{softComPrec}$ scores upwards of 
0.9), but these allow many additional analyses (hence the scores below
0.6).

\section{Conclusion}

We have presented a framework for simple dependency annotation that overcomes some of the representational limitations of 
unlabeled dependency grammar and embraces the practical realities of resource-building efforts.
Pilot studies (in multiple languages and domains, supported by a human-readable notation and a suite of open-source tools)
showed this approach lends itself to rapid annotation with minimal training.

The next step will be to develop algorithms exploiting these representations for learning parsers. 
Other future extensions might include additional expressive mechanisms (e.g., multi-headedness, labels), 
crowdsourcing of FUDG annotations \cite{snow-08},\footnote{We ran a pilot survey on Mechanical Turk and found 80 workers with experience in sentence diagramming (\fnref{sentdiag}), 
suggesting that simple syntactic annotations may be feasibly produced by non-expert annotators.}
or even a semantic counterpart to the syntactic representation.

{\small
\section*{Acknowledgments}
We > thank < ([Lukas~Biewald], 
[Yoav~Goldberg], 
[Kyle~Jerro],
[Vijay~John], 
[Lori~Levin], 
[Andr\'{e}~Martins], 
and (several anonymous reviewers*))
for < (their > insights).
This research was supported in part by the U.~S.~Army Research Laboratory and the
U.~S.~Army Research Office under contract/grant number
W911NF-10-1-0533 and by NSF grant IIS-1054319.

}

\bibliographystyle{acl}
\bibliography{fudg_tr}

\begin{thebibliography}{}

\bibitem[\protect\citename{Baldwin and Kim}2010]{baldwin-10}
Timothy Baldwin and Su~Nam Kim.
\newblock 2010.
\newblock Multiword expressions.
\newblock In Nitin Indurkhya and Fred~J. Damerau, editors, {\em Handbook of
  Natural Language Processing, Second Edition}. {CRC} Press, Taylor and Francis
  Group, Boca Raton, {FL}.

\bibitem[\protect\citename{Bergen and Chang}2005]{bergen-05}
Benjamin~K. Bergen and Nancy Chang.
\newblock 2005.
\newblock Embodied {C}onstruction {G}rammar in simulation-based language
  understanding.
\newblock In {Jan-Ola} \"{O}stman and Mirjam Fried, editors, {\em Construction
  grammars: cognitive grounding and theoretical extensions}, pages 147--190.
  John Benjamins, Amsterdam.

\bibitem[\protect\citename{Bies \bgroup et al.\egroup }2012]{webtb}
Ann Bies, Justin Mott, Colin Warner, and Seth Kulick.
\newblock 2012.
\newblock English {W}eb {T}reebank.
\newblock Technical Report {LDC2012T13}, Linguistic Data Consortium,
  Philadelphia, {PA}.

\bibitem[\protect\citename{B\"{o}hmov\'{a} \bgroup et al.\egroup
  }2003]{bohmova-03}
Alena B\"{o}hmov\'{a}, Jan Haji\v{c}, Eva Haji\v{c}ov\'{a}, Barbora Hladk\'{a},
  and Anne Abeill\'{e}.
\newblock 2003.
\newblock The {P}rague {D}ependency {T}reebank: a three-level annotation
  scenario.
\newblock In {\em Treebanks: building and using parsed corpora}, pages
  103--127. Springer.

\bibitem[\protect\citename{Buchholz and Marsi}2006]{buchholz-06}
Sabine Buchholz and Erwin Marsi.
\newblock 2006.
\newblock {CoNLL-X} shared task on multilingual dependency parsing.
\newblock In {\em Proceedings of the Tenth Conference on Computational Natural
  Language Learning ({CoNLL-X})}, pages 149--164, New York City, June.
  Association for Computational Linguistics.

\bibitem[\protect\citename{Chaiken and Kleitman}1978]{chaiken-78}
Seth Chaiken and Daniel~J. Kleitman.
\newblock 1978.
\newblock Matrix {T}ree {T}heorems.
\newblock {\em Journal of Combinatorial Theory, Series A}, 24(3):377--381, May.

\bibitem[\protect\citename{Chomsky}1957]{chomsky-57}
Noam Chomsky.
\newblock 1957.
\newblock {\em Syntactic Structures}.
\newblock Mouton, La Haye.

\bibitem[\protect\citename{Clark and Curran}2006]{clark-06}
Stephen Clark and James Curran.
\newblock 2006.
\newblock Partial training for a lexicalized-grammar parser.
\newblock In {\em Proceedings of the Human Language Technology Conference of
  the {NAACL} ({HLT-NAACL} 2006)}, pages 144--151, New York City, {USA}, June.
  Association for Computational Linguistics.

\bibitem[\protect\citename{Croft}2001]{croft-01}
William Croft.
\newblock 2001.
\newblock {\em Radical {C}onstruction {G}rammar: Syntactic Theory in
  Typological Perspective}.
\newblock Oxford University Press, Oxford.

\bibitem[\protect\citename{de Marneffe and Manning}2008]{de_marneffe-08}
{Marie-Catherine} de~Marneffe and Christopher~D. Manning.
\newblock 2008.
\newblock Stanford typed dependencies manual.
\newblock \url{http://nlp.stanford.edu/downloads/dependencies\_manual.pdf}.

\bibitem[\protect\citename{Florey}2006]{florey-06}
Kitty~Burns Florey.
\newblock 2006.
\newblock {\em Sister {B}ernadette's Barking Dog: The quirky history and lost
  art of diagramming sentences}.
\newblock Melville House, New York, October.

\bibitem[\protect\citename{Fossum and Levy}2012]{fossum-12}
Victoria Fossum and Roger Levy.
\newblock 2012.
\newblock Sequential vs. hierarchical syntactic models of human incremental
  sentence processing.
\newblock In {\em Proceedings of the 3rd Workshop on Cognitive Modeling and
  Computational Linguistics ({CMCL} 2012)}, pages 61--69, Montr\'{e}al, Canada,
  June. Association for Computational Linguistics.

\bibitem[\protect\citename{Foster \bgroup et al.\egroup }2011]{foster-11}
Jennifer Foster, Ozlem Cetinoglu, Joachim Wagner, Joseph Le~Roux, Stephen
  Hogan, Joakim Nivre, Deirdre Hogan, and Josef van Genabith.
\newblock 2011.
\newblock \#hardtoparse: {POS} {T}agging and {P}arsing the {T}witterverse.
\newblock In {\em Proceedings of the 2011 {AAAI} Workshop on Analyzing
  Microtext}, pages 20--25, San Francisco, {CA}, August. {AAAI} Press.

\bibitem[\protect\citename{Group}2007]{pdtb-manual}
The {PDTB}~Research Group.
\newblock 2007.
\newblock The {P}enn {D}iscourse {T}reebank 2.0 annotation manual.
\newblock Technical Report {IRCS-08-01}, Institute for Research in Cognitive
  Science, University of Pennsylvania, Philadelphia, {PA}, December.

\bibitem[\protect\citename{Haji\v{c} \bgroup et al.\egroup }2001]{hajic-01}
Jan Haji\v{c}, Barbora~Vidov\'{a} Hladk\'{a}, and Petr Pajas.
\newblock 2001.
\newblock The {P}rague {D}ependency {T}reebank: annotation structure and
  support.
\newblock In {\em Proceedings of the {IRCS} Workshop on Linguistic Databases},
  pages 105--114. University of Pennsylvania, Philadelphia, {USA}.

\bibitem[\protect\citename{Haji\v{c}}1998]{hajic-98}
Jan Haji\v{c}.
\newblock 1998.
\newblock Building a syntactically annotated corpus: the {P}rague {D}ependency
  {T}reebank.
\newblock In Eva Haji\v{c}ov\'{a}, editor, {\em Issues of Valency and Meaning.
  Studies in Honor of Jarmila Panevov\'{a}}, pages 12--19. Prague Karolinum,
  Charles University Press, Prague.

\bibitem[\protect\citename{Hudson}1984]{hudson-84}
Richard~A. Hudson.
\newblock 1984.
\newblock {\em Word {G}rammar}.
\newblock Blackwell, Oxford.

\bibitem[\protect\citename{Hwa}1999]{hwa-99}
Rebecca Hwa.
\newblock 1999.
\newblock Supervised grammar induction using training data with limited
  constituent information.
\newblock In {\em Proceedings of the 37th Annual Meeting of the Association for
  Computational Linguistics ({ACL-99})}, pages 73--79, College Park, Maryland,
  {USA}, June. Association for Computational Linguistics.

\bibitem[\protect\citename{Johansson and Nugues}2007]{johansson-07}
Richard Johansson and Pierre Nugues.
\newblock 2007.
\newblock Extended constituent-to-dependency conversion for {E}nglish.
\newblock In Joakim Nivre, {Heiki-Jaan} Kaalep, Kadri Muischnek, and Mare Koit,
  editors, {\em Proceedings of the 16th Nordic Conference of Computational
  Linguistics ({NODALIDA-2007})}, pages 105--112, Tartu, Estonia, May.

\bibitem[\protect\citename{Kolln and Funk}1994]{kolln-94}
Martha Kolln and Robert Funk.
\newblock 1994.
\newblock {\em Understanding {E}nglish Grammar}.
\newblock Macmillan, New York.

\bibitem[\protect\citename{K\"{u}bler \bgroup et al.\egroup }2009]{kubler-09}
Sandra K\"{u}bler, Ryan {McDonald}, and Joakim Nivre.
\newblock 2009.
\newblock {\em Dependency {P}arsing}.
\newblock Number~2 in Synthesis Lectures on Human Language Technologies. Morgan
  \& Claypool, San Rafael, {CA}, January.

\bibitem[\protect\citename{Kuhlmann and Nivre}2006]{kuhlmann-06}
Marco Kuhlmann and Joakim Nivre.
\newblock 2006.
\newblock Mildly non-projective dependency structures.
\newblock In {\em Proceedings of the {COLING}/{ACL} 2006 Main Conference Poster
  Sessions}, pages 507--514, Sydney, Australia, July. Association for
  Computational Linguistics.

\bibitem[\protect\citename{Marcus \bgroup et al.\egroup }1993]{ptb}
Mitchell~P. Marcus, Beatrice Santorini, and Mary~Ann Marcinkiewicz.
\newblock 1993.
\newblock Building a large annotated corpus of {E}nglish: the {P}enn
  {T}reebank.
\newblock {\em Computational Linguistics}, 19(2):313--330.

\bibitem[\protect\citename{Mare{\v{c}}ek \bgroup et al.\egroup
  }2013]{marecek-13}
David Mare{\v{c}}ek, Martin Popel, Loganathan Ramasamy, Jan
  {\v{S}}t{\v{e}}p{\'{a}}nek, Daniel Zeman, Zden{\v{e}}k
  {\v{Z}}abokrtsk{\'{y}}, and Jan Haji{\v{c}}.
\newblock 2013.
\newblock Cross-language study on influence of coordination style on dependency
  parsing performance.
\newblock Technial Report~49, {\'{U}}{FAL} {MFF} {UK}, January.

\bibitem[\protect\citename{Margoliash}2010]{margoliash-10}
Jonathan Margoliash.
\newblock 2010.
\newblock {Matrix-Tree} {T}heorem for directed graphs.
\newblock
  \url{http://www.math.uchicago.edu/~may/VIGRE/VIGRE2010/REUPapers/Margoliash.%
pdf}, August.

\bibitem[\protect\citename{Mel'\v{c}uk}1988]{melcuk-88}
Igor~Aleksandrovi\v{c} Mel'\v{c}uk.
\newblock 1988.
\newblock {\em Dependency Syntax: Theory and Practice}.
\newblock {SUNY} Press, Albany, {NY}.

\bibitem[\protect\citename{Nivre}2005]{nivre-05}
Joakim Nivre.
\newblock 2005.
\newblock Dependency grammar and dependency parsing.
\newblock Technical Report {MSI} report 05133, V\"{a}xj\"{o} University School
  of Mathematics and Systems Engineering, V\"{a}xj\"{o}, Sweden.

\bibitem[\protect\citename{Osborne \bgroup et al.\egroup }2012]{osborne-12}
Timothy Osborne, Michael Putnam, and Thomas Gro{\ss}.
\newblock 2012.
\newblock Catenae: introducing a novel unit of syntactic analysis.
\newblock {\em Syntax}, 15(4):354--396.

\bibitem[\protect\citename{Owoputi \bgroup et al.\egroup }2012]{owoputi-tr}
Olutobi Owoputi, Brendan O'Connor, Chris Dyer, Kevin Gimpel, and Nathan
  Schneider.
\newblock 2012.
\newblock Part-of-speech tagging for {T}witter: Word clusters and other
  advances.
\newblock Technical Report CMU-ML-12-107, Carnegie Mellon University,
  Pittsburgh, Pennsylvania, September.

\bibitem[\protect\citename{Reed and Kellogg}1877]{reed-77}
Alonzo Reed and Brainerd Kellogg.
\newblock 1877.
\newblock {\em Work on {E}nglish grammar \& composition}.
\newblock Clark \& Maynard.

\bibitem[\protect\citename{Sangati and Mazza}2009]{sangati-09}
Federico Sangati and Chiara Mazza.
\newblock 2009.
\newblock An {E}nglish dependency treebank \`{a} la {T}esni\`{e}re.
\newblock In Marco Passarotti, Adam Przepi\'{o}rkowski, Savina Raynaud, and
  Frank Van~Eynde, editors, {\em Proceedings of the Eigth International
  Workshop on Treebanks and Linguistic Theories}, pages 173--184, Milan, Italy,
  December. {EDUCatt}.

\bibitem[\protect\citename{Schneider \bgroup et al.\egroup }2013]{fudg-law}
Nathan Schneider, Brendan O'Connor, Naomi Saphra, David Bamman, Manaal Faruqui,
  Noah~A. Smith, Chris Dyer, and Jason Baldridge.
\newblock 2013.
\newblock A framework for (under)specifying dependency syntax without
  overloading annotators.
\newblock In {\em Proceedings of the 7th Linguistic Annotation Workshop \&
  Interoperability with Discourse}, Atlanta, Georgia, {USA}, August.
  Association for Computational Linguistics.

\bibitem[\protect\citename{Schneider}1998]{schneider-98}
Gerold Schneider.
\newblock 1998.
\newblock {\em A linguistic comparison of constituency, dependency and link
  grammar}.
\newblock Master's thesis, University of Zurich, July.

\bibitem[\protect\citename{Sgall \bgroup et al.\egroup }1986]{sgall-86}
Petr Sgall, Eva Haji\v{c}ov\'{a}, and Jarmila Panevov\'{a}.
\newblock 1986.
\newblock {\em The Meaning of the Sentence in its Semantic and Pragmatic
  Aspects}.
\newblock Reidel, Dordrecht and Academia, Prague, May.

\bibitem[\protect\citename{Shieber}1992]{shieber-92}
Stuart~M. Shieber.
\newblock 1992.
\newblock {\em Constraint-Based Grammar Formalisms}.
\newblock {MIT} Press, Cambridge, {MA}.

\bibitem[\protect\citename{Sleator and Temperly}1993]{sleator-93}
Daniel Sleator and Davy Temperly.
\newblock 1993.
\newblock Parsing {E}nglish with a link grammar.
\newblock In {\em Proceedings of the Third International Workshop on Parsing
  Technology ({IWPT'93})}, pages 277--292, Tilburg, Netherlands, August.

\bibitem[\protect\citename{Smith and Smith}2007]{smith-07}
David~A. Smith and Noah~A. Smith.
\newblock 2007.
\newblock Probabilistic models of nonprojective dependency trees.
\newblock In {\em Proceedings of the 2007 Joint Conference on Empirical Methods
  in Natural Language Processing and Computational Natural Language Learning
  ({EMNLP-CoNLL} 2007)}, pages 132--140, Prague, Czech Republic, June.
  Association for Computational Linguistics.

\bibitem[\protect\citename{Snow \bgroup et al.\egroup }2008]{snow-08}
Rion Snow, Brendan {O'Connor}, Daniel Jurafsky, and Andrew Ng.
\newblock 2008.
\newblock Cheap and fast --- but is it good? {E}valuating non-expert
  annotations for natural language tasks.
\newblock In {\em Proceedings of the 2008 Conference on Empirical Methods in
  Natural Language Processing ({EMNLP} 2008)}, pages 254--263, Honolulu,
  Hawaii, October. Association for Computational Linguistics.

\bibitem[\protect\citename{Steedman}2000]{steedman-00}
Mark Steedman.
\newblock 2000.
\newblock {\em The Syntatic Process}.
\newblock {MIT} Press, Cambridge, {MA}.

\bibitem[\protect\citename{Steels \bgroup et al.\egroup }2011]{steels-11}
Luc Steels, {Jan-Ola} \"{O}stman, and Kyoko Ohara, editors.
\newblock 2011.
\newblock {\em Design patterns in {F}luid {C}onstruction {G}rammar}.
\newblock Number~11 in Constructional Approaches to Language. John Benjamins,
  Amsterdam.

\bibitem[\protect\citename{Tesni\`{e}re}1959]{tesniere-59}
Lucien Tesni\`{e}re.
\newblock 1959.
\newblock {\em El\'{e}ments de Syntaxe Structurale}.
\newblock Klincksieck, Paris.

\bibitem[\protect\citename{Tomasello}2003]{tomasello-03}
Michael Tomasello.
\newblock 2003.
\newblock {\em Constructing a Language: A Usage-Based Theory of Language
  Acquisition}.
\newblock Harvard University Press, Cambridge, {MA}.

\bibitem[\protect\citename{Uno}1996]{uno-96}
Takeaki Uno.
\newblock 1996.
\newblock An algorithm for enumerating all directed spanning trees in a
  directed graph.
\newblock In Tetsuo Asano, Yoshihide Igarashi, Hiroshi Nagamochi, Satoru
  Miyano, and Subhash Suri, editors, {\em Algorithms and Computation}, number
  1178 in Lecture Notes in Computer Science, pages 166--173. Springer, Berlin,
  January.

\end{thebibliography}

\end{document}